\documentclass[runningheads]{llncs}
\usepackage{graphicx}
\usepackage{comment}
\usepackage{amsmath,amssymb} 
\usepackage{color}
\usepackage{subfigure}
\usepackage{array}
\usepackage{booktabs}
\usepackage{colortbl}
\usepackage{arydshln}
\usepackage{verbatim} 
\usepackage{gensymb} 
\usepackage{multirow}
\usepackage{tabu}  
\usepackage{adjustbox}
\usepackage{xcolor}
\usepackage{lipsum}
\usepackage{siunitx}
\usepackage{pifont}

\usepackage[pagebackref=true,breaklinks=true,letterpaper=true,colorlinks,bookmarks=false]{hyperref}

\begin{document}

\newcommand{\fullname}{Virtual Image Dataset for Illumination Transfer}
\newcommand{\name}{VIDIT}

\pagestyle{headings}
\mainmatter
\def\ECCVSubNumber{271}  

\newcommand\blfootnote[1]{%
	\begingroup
	\renewcommand\thefootnote{}\footnote{#1}%
	\addtocounter{footnote}{-1}%
	\endgroup
}

\title{VIDIT: Virtual Image Dataset \\for Illumination Transfer}

\titlerunning{VIDIT}
%
\author{\author{Majed El Helou \and
Ruofan Zhou \and
Johan Barthas \and
Sabine S\"usstrunk}
\authorrunning{M. El Helou et al.}
%
\institute{School of Computer and Communication Sciences, EPFL, Switzerland \\
\email{\{majed.elhelou,ruofan.zhou\}@epfl.ch}}}
\maketitle

\begin{abstract}
Deep image relighting is gaining more interest lately, as it allows photo enhancement through illumination-specific retouching without human effort. Aside from aesthetic enhancement and photo montage, image relighting is valuable for domain adaptation, whether to augment datasets for training or to normalize input test data. Accurate relighting is, however, very challenging for various reasons, such as the difficulty in removing and recasting shadows and the modeling of different surfaces.

We present a novel dataset, the Virtual Image Dataset for Illumination Transfer (VIDIT), in an effort to create a reference evaluation benchmark and to push forward the development of illumination manipulation methods. Virtual datasets are not only an important step towards achieving real-image performance but have also proven capable of improving training even when real datasets are possible to acquire and available. VIDIT contains 300 virtual scenes used for training, where every scene is captured 40 times in total: from 8 equally-spaced azimuthal angles, each lit with 5 different illuminants. \url{https://github.com/majedelhelou/VIDIT}

\end{abstract}

\newcommand{\teaser}[1]{\includegraphics[width=0.25\linewidth,clip]{#1}}
\begin{figure}[t]
    \centering
    \begin{tabu}{cccc}
        \teaser{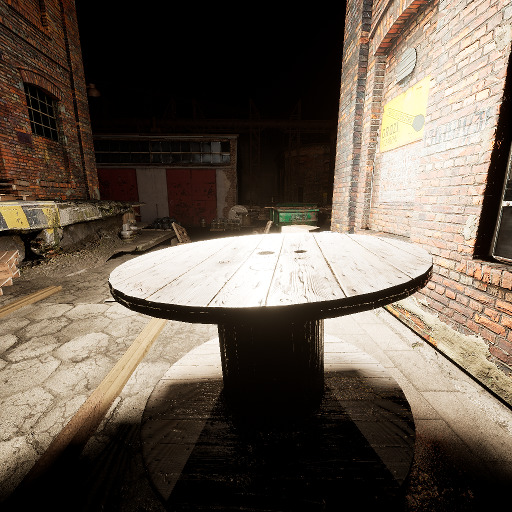}&
        \teaser{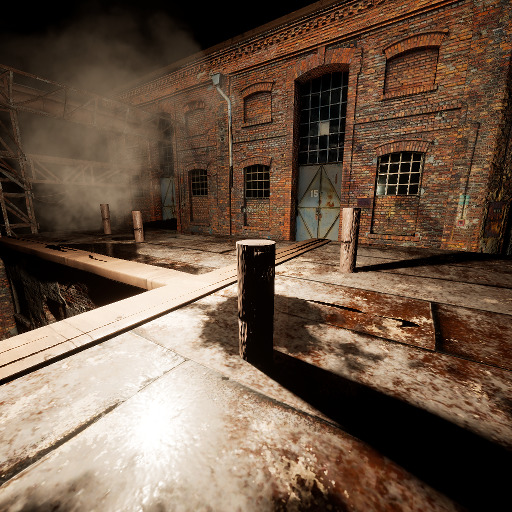}&
        \teaser{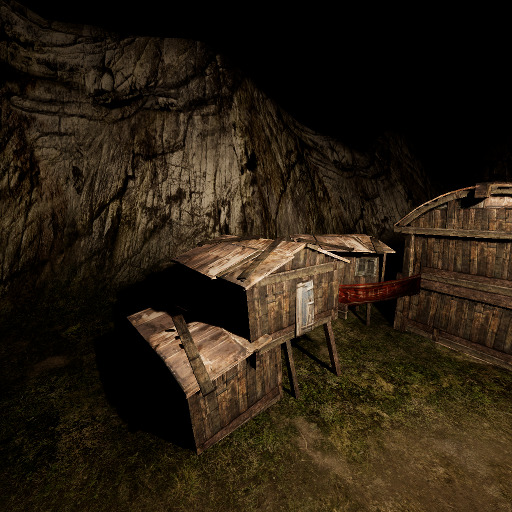}&
        \teaser{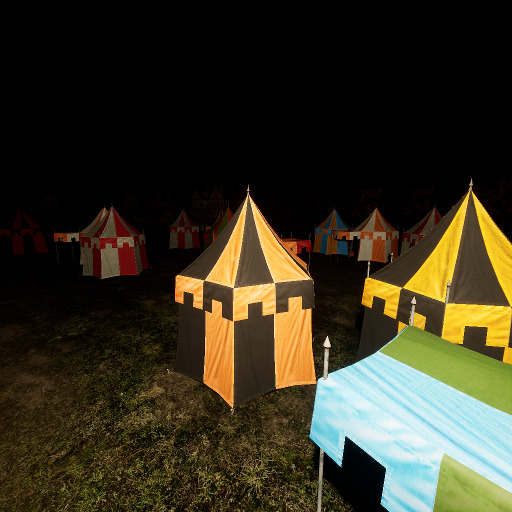}\\
        \teaser{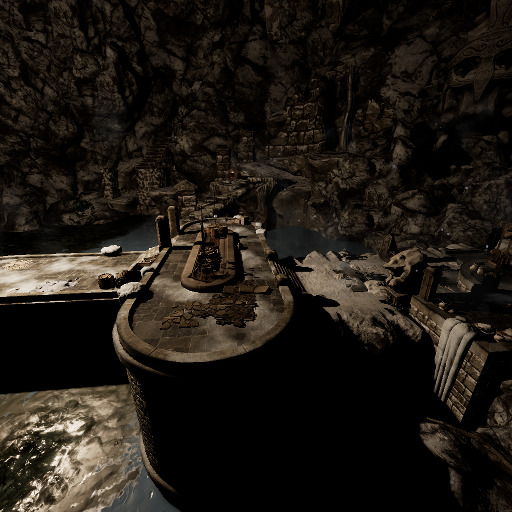}&
        \teaser{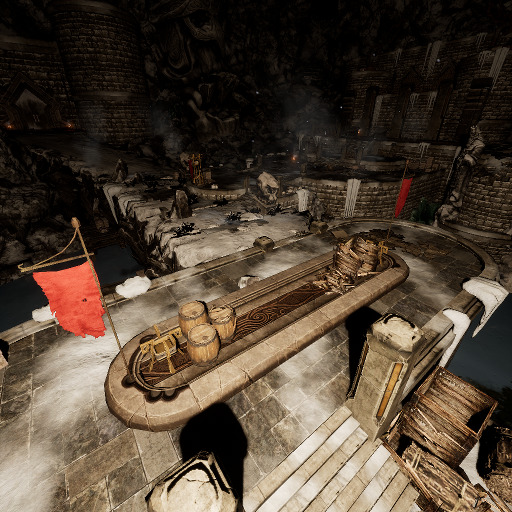}&
        \teaser{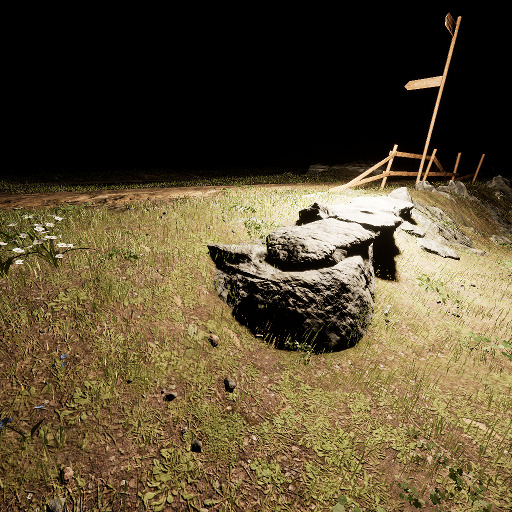}&
        \teaser{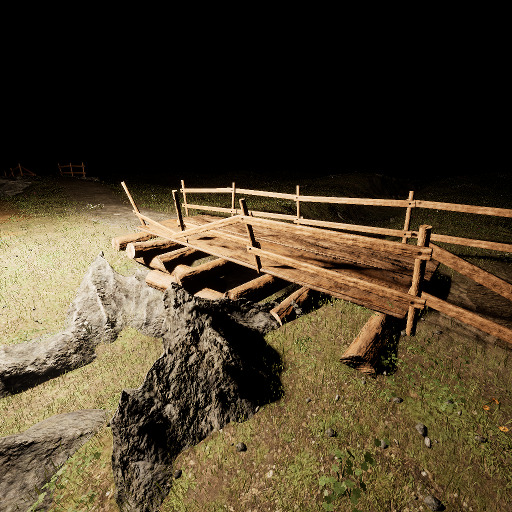}
    \end{tabu}
    \caption{A selection of scenes captured at 4500K, with the light source placed at (N,NW,SE,S,W,S,NE,W) from left to right then top down. Images are shown downsampled by 2.}
    \label{fig:teaser}
\end{figure}

\section{Dataset Description}
\label{sec:introduction}

\subsection{Content}
VIDIT includes in total 390 different scenes, each of which is captured with 40 predetermined illumination settings, resulting in 15,600 images. The illumination settings are all the combinations of 5 color temperatures (2500K, 3500K, 4500K, 5500K and 6500K) and 8 light directions (N, NE, E, SE, S, SW, W, NW), each of which is shown in Fig.~\ref{fig:teaser_full} for a sample image from our dataset. The directions are equally-spaced azimuthal angles, with a fixed polar angle. The dataset includes both indoor and outdoor scenes, and miscellaneous objects with different surfaces and materials. All scenes are rendered with a resolution of $1024 \times 1024$ pixels, contain different materials such as metal, wood, stone, water, plant, fabric, smoke, fire, plastic, etc. The illumination settings as well as the depth information are recorded with each rendered images.

VIDIT is split into different mutually-exclusive sets; train (300 scenes), validation (45 scenes), and test (45 scenes). The test set is kept private for benchmarking purposes, while the train and validation sets are made public for both training and method evaluation.

\begin{table}[]
\centering
\begin{tabular}{cccccc}
\toprule
           & \multirow{2}{*}{Scenes} & Light  & Color & \multirow{2}{*}{Total} & \multirow{2}{*}{Availability} \\
           & & Orientations & Temperatures & &  \\ \cline{2-6}
Train      & 300    & 8                  & 5                  & 12,000 & Public       \\
Validation & 45     & 8                  & 5                  & 1,800  & Public       \\
Test       & 45     & 8                  & 5                  & 1,800  & Private     \\ 
\bottomrule
\end{tabular}
\vspace{.2cm}
\caption{VIDIT content distribution. The dataset is split into train, validation, and test, the latter kept private for benchmarking. Each scene is captured under 40 different illumination settings, and is rendered with a $1024 \times 1024$ pixel resolution.}
\end{table}

\subsection{Acquisition}
\begin{figure}
    \centering
    \includegraphics[width=0.8\linewidth]{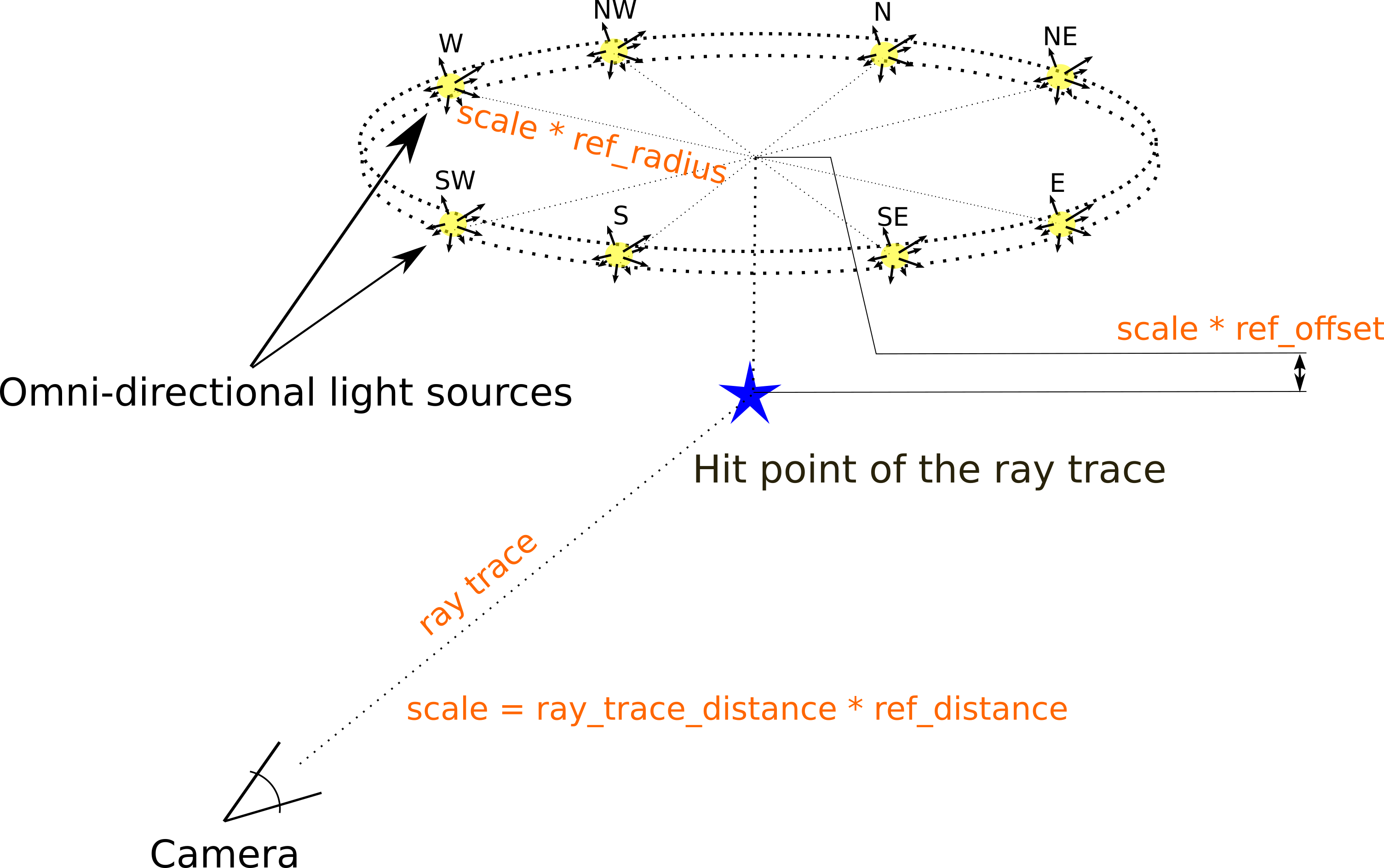}
    \caption{We place a set of light sources above the hit point of the ray trace to keep the same light settings for each of our captured scenes. We automatically adjust to the environment's scale by adapting our reference radius, reference offset, and reference camera distance. During capture, we turn on the light source corresponding to the selected cardinal direction (N, NE, E, SE, S, SW, W, NW) and turn off all the others. We also vary the light source's color temperature for 5 different captures.}
    \label{fig:scene_capture}
\end{figure}

We use Unreal Engine 4 to render high-resolution and realistic scenes to build our dataset. The scenes are obtained from a variety of different virtual environments, which are scaled into a uniform reference space before running the illumination and rendering process. The selection of scenes (or fields of view) is done manually to avoid any duplicates or very similar content being captured accidentally by two images. The scene is illuminated with an omni-directional light source, with 5 color temperatures, and the source is positioned at the 8 different azimuthal angles illustrated in Fig.~\ref{fig:scene_capture}. The fields of view with walls or large obstructive objects that do not permit proper illumination from all angles were manually removed from the dataset.

\newpage

\section{Related Work}
We list a series of recent work related to our dataset which we found relevant.
\\

\noindent \textbf{A Dataset of Multi-Illumination Images in the Wild} ICCV'19~\cite{murmann2019dataset}. \\
This paper presents a dataset of \textit{interior} scenes only, mostly objects in small and controlled environments. Each scene is captured with a flash from 25 different orientations. The illuminant itself, however, does not vary across the dataset. \\

\noindent \textbf{Underexposed Photo Enhancement Using Deep Illumination Estimation} CVPR'19~\cite{wang2019underexposed}. \\
This dataset contains pairs of underexposed images. In opposition to the MIT-Adobe FiveK dataset, which is geared towards general photo enhancement, the authors focus specifically on enhancing underexposed photos. The ground-truth enhanced images are created by photo experts using Adobe Lightroom. \\

\noindent \textbf{Deep Image-Based Relighting from Optimal Sparse Samples} TOG'18~\cite{xu2018deep}. \\
This paper, similar to ~\cite{murmann2019dataset}, also proposes a dataset of scenes with different light directions. The differences are that the images are rendered and that the light directions are randomized. \\

\noindent \textbf{Single Image Portrait Relighting} TOG'19~\cite{sun2019single}. \\
This paper aims at relighting human portrait photos. The method is trained on a dataset consisting of 18 individuals captured under different directional light sources. The capture is under controlled settings, with the individual illuminated by a sphere with numerous lights. Specifically targeted at face relighting, the method does not extend to general scenes. \\

\noindent \textbf{Deep Face Normalization} TOG'19~\cite{nagano2019deep}. \\
This paper partly addresses relighting. It aims at normalizing faces in portrait images, i.e., removing distortions and non-neutral facial expressions, and relighting to simulate an evenly-lit environment. It does not address relighting to different light sources or to a desired light direction. \\


There is a rich recent body of literature on style transfer methods and on intrinsic image decomposition. To name a few, IIW~\cite{bell2014intrinsic} and SAW~\cite{kovacs2017shading} contain human-labeled reflectance and shading annotations, BigTime~\cite{li2018learning} contains time-lapse data where the scenes are illuminated under varying lighting conditions. These works, however, do not provide pairs of scenes captured with the same lighting conditions (color temperature, direction) that allow a quantitative evaluation of illumination transfer from one photo to another as our dataset does.

\section{Discussion and Applications}
Our dataset can be used in different applications, both for learning and evaluation. It can be used to predict the orientation of the illuminant in a scene, as well as its color temperature. More interestingly, it can be employed in domain adaptation to transform any input image such that it matches the target domain's illumination settings (which is an any-to-one illumination style transfer problem). An even more general and challenging application is what we call the any-to-any illumination style transfer, which consists of mapping an input image with certain illumination settings to match the illumination settings of another input guide image. 
Both the any-to-one and any-to-any problems can be addressed with a guide input image from the target domain, or with input illumination settings (orientation and color temperature). And all applications can be evaluated with objective metrics, since our dataset includes all of the different illumination settings for every single captured scene. 

Some use cases of image relighting methods would be
\begin{itemize}
    \item \textbf{Domain adaptation for data augmentation}: input scenes can be transformed into a variety of illumination settings and added to training datasets to improve the robustness of deep networks to light variation.
    \item \textbf{Domain adaptation for normalization}: input scenes can be transformed from many illumination settings into a single reference one, before they are fed into a pre-trained deep network or a search and retrieval method.
    \item \textbf{Photo editing and shadow manipulation}: a scene can be relit for aesthetic reasons, to match the desired result of the user.
    \item \textbf{Photo montage}: combining multiple images into one necessitates that they all match the same final illumination settings. This can be achieved using an image relighting method.
\end{itemize}

\newcommand{\teaserimg}[1]{\includegraphics[width=0.2\linewidth,clip]{#1}}
\begin{figure}[t]
    \centering
    \begin{tabu}{ccccc}
        
        \teaserimg{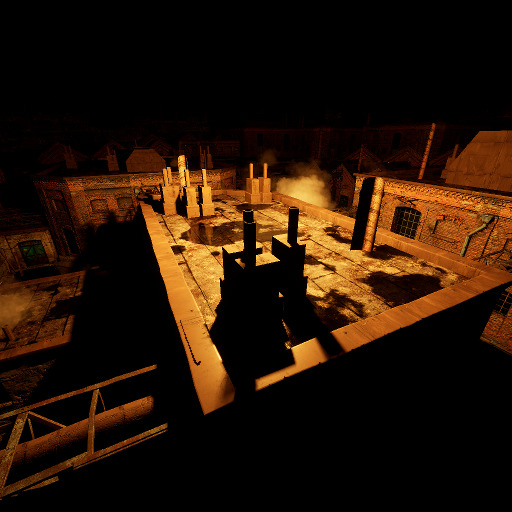}&
        \teaserimg{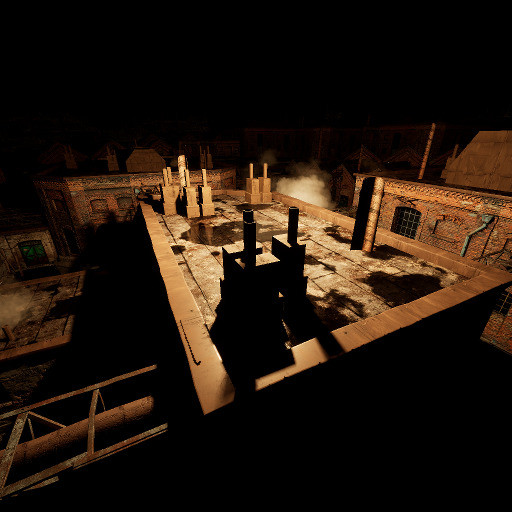}&
        \teaserimg{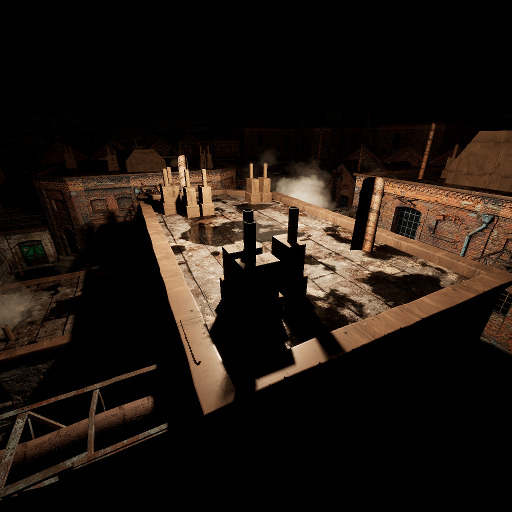}&
        \teaserimg{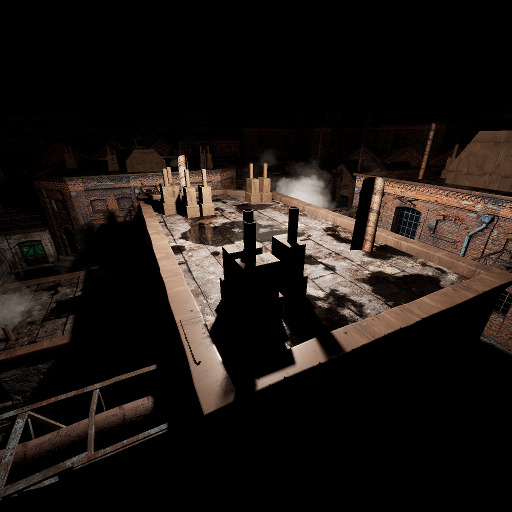}&
        \teaserimg{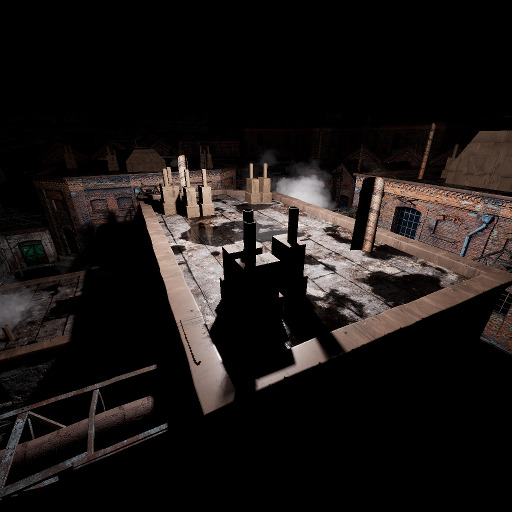}\\
        \teaserimg{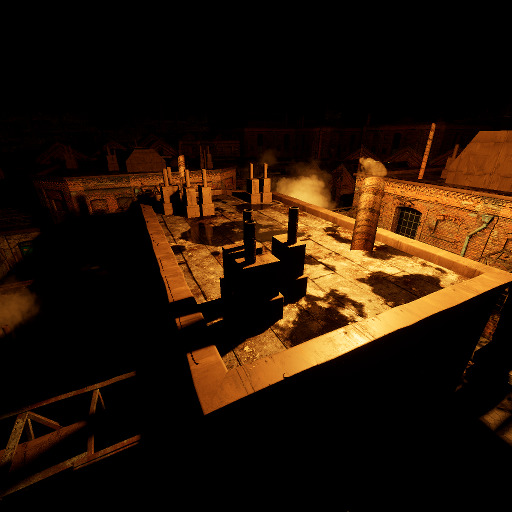}&
        \teaserimg{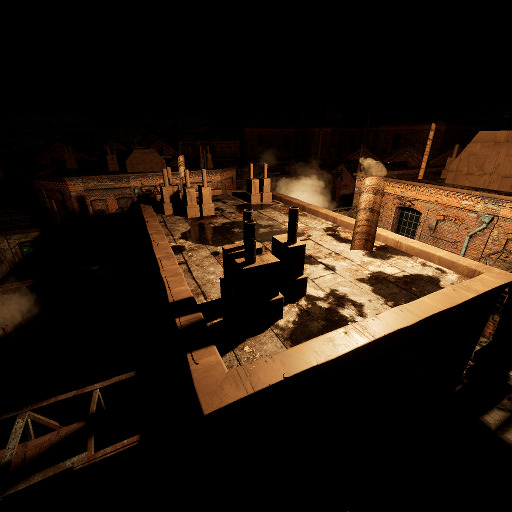}&
        \teaserimg{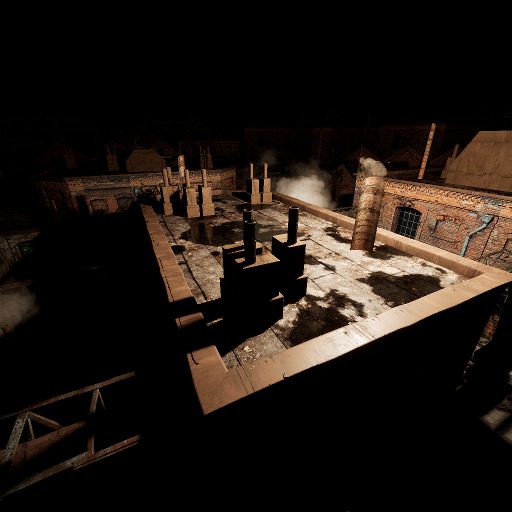}&
        \teaserimg{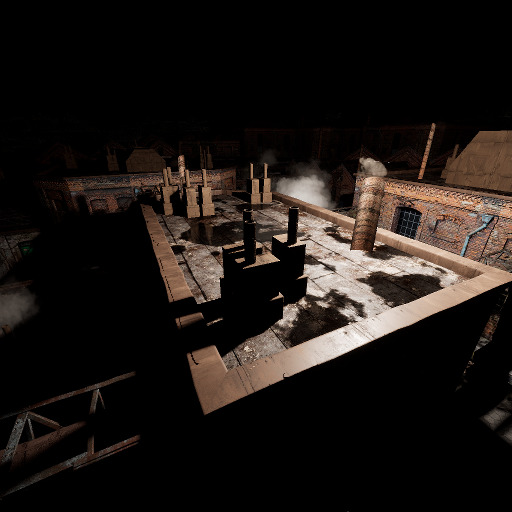}&
        \teaserimg{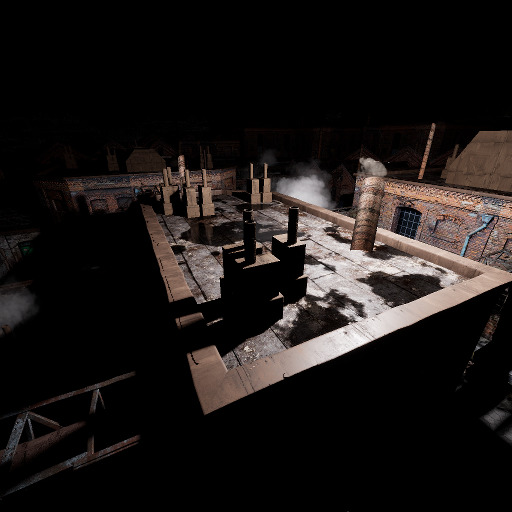}\\
        \teaserimg{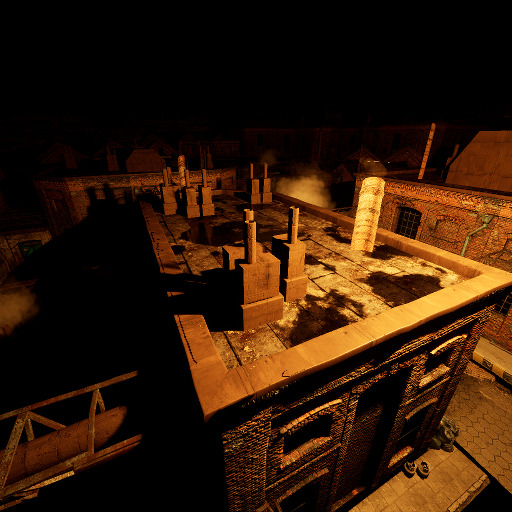}&
        \teaserimg{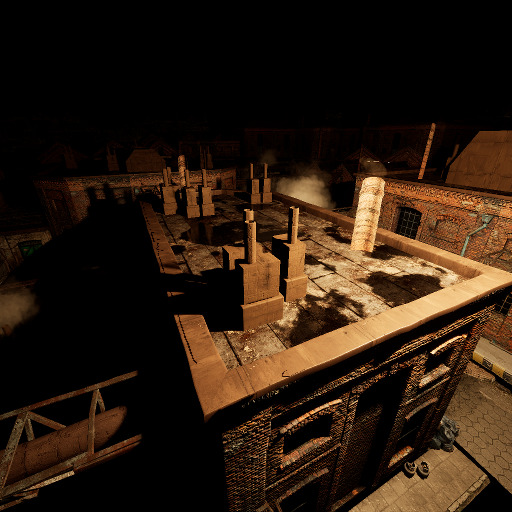}&
        \teaserimg{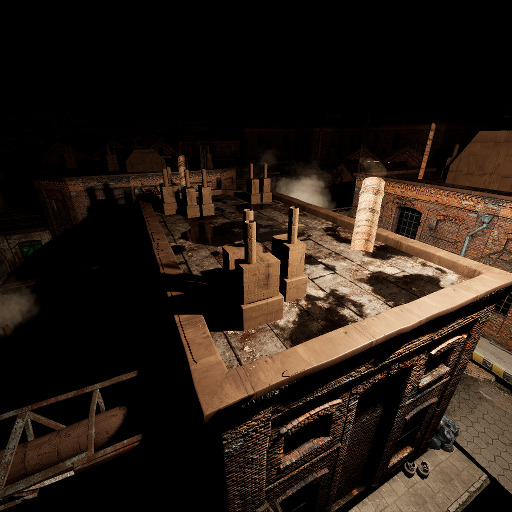}&
        \teaserimg{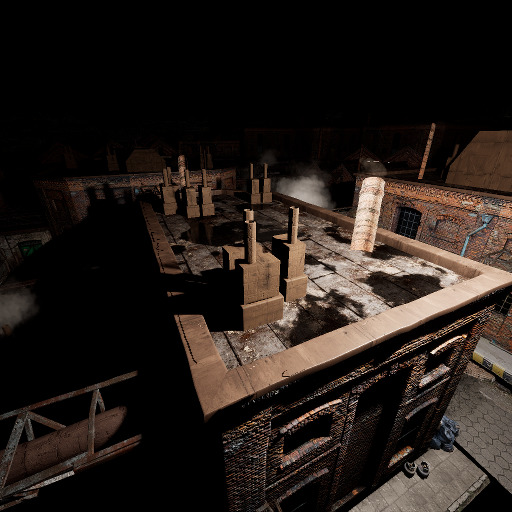}&
        \teaserimg{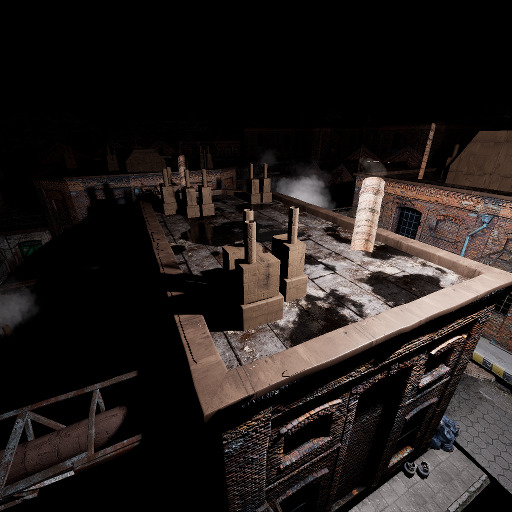}\\
        \teaserimg{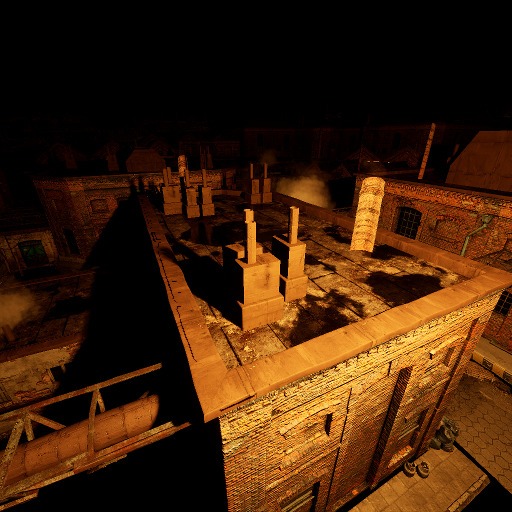}&
        \teaserimg{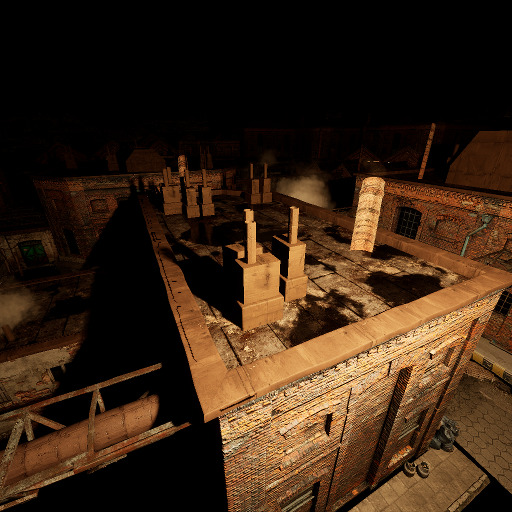}&
        \teaserimg{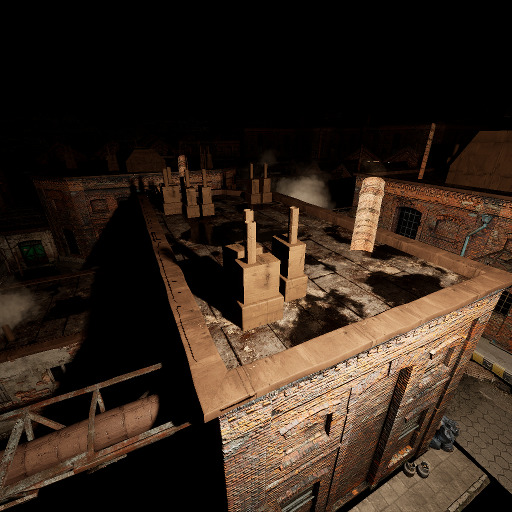}&
        \teaserimg{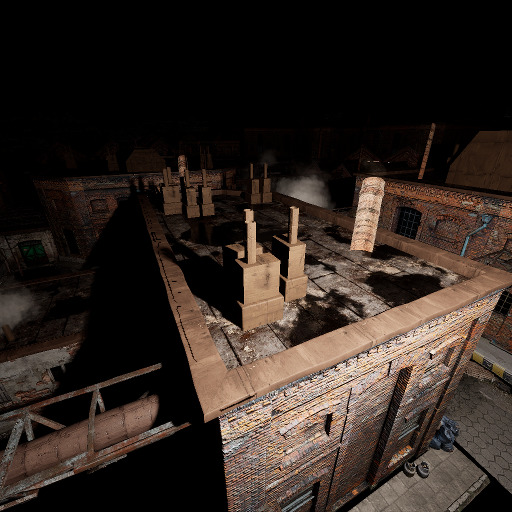}&
        \teaserimg{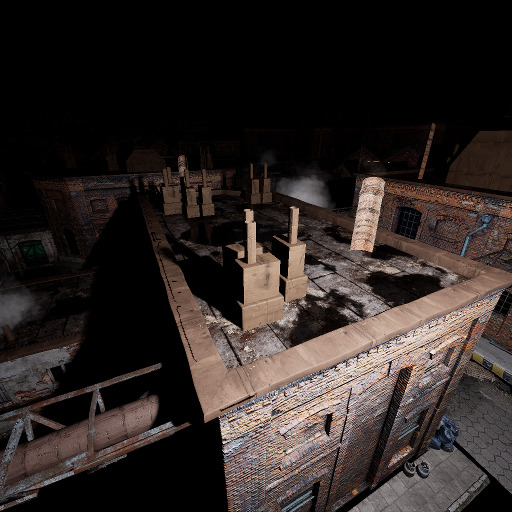}\\
        \teaserimg{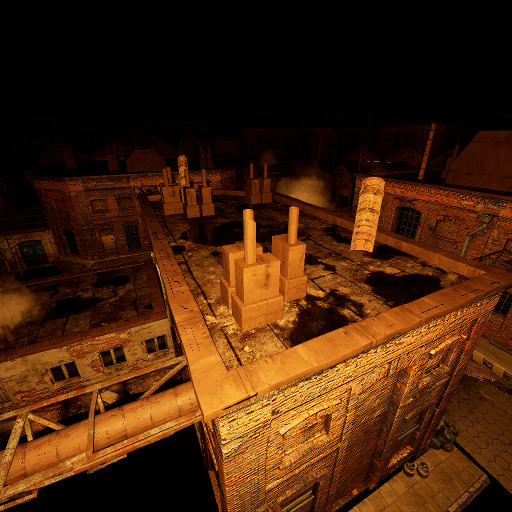}&
        \teaserimg{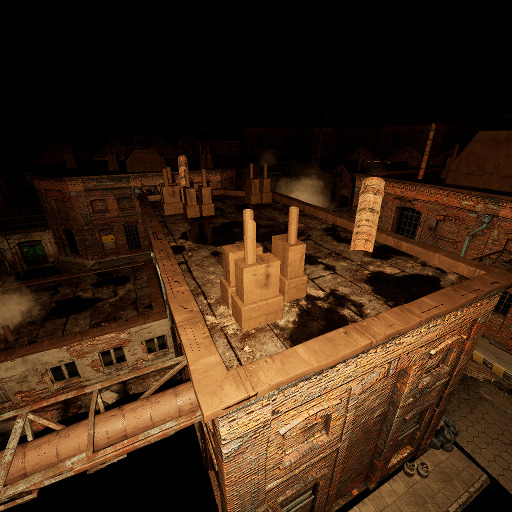}&
        \teaserimg{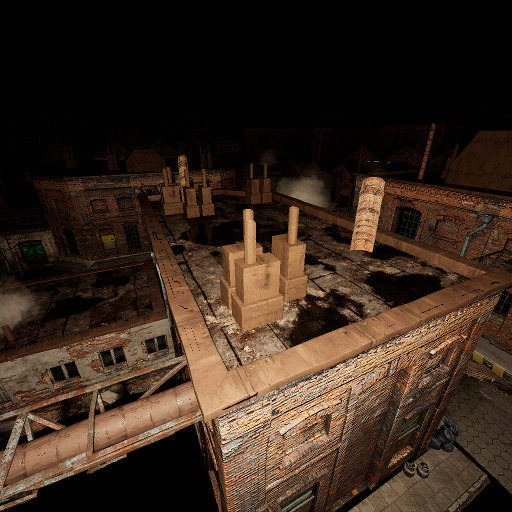}&
        \teaserimg{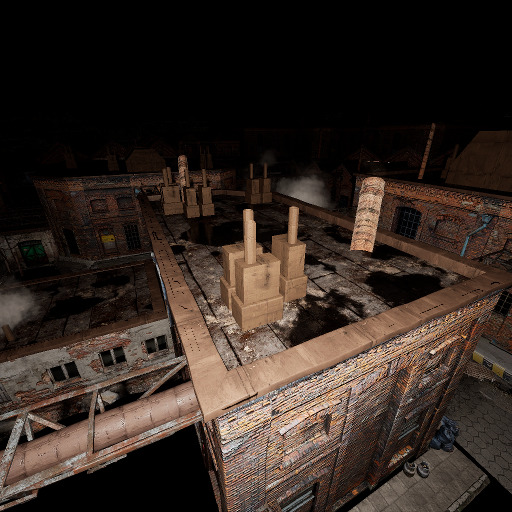}&
        \teaserimg{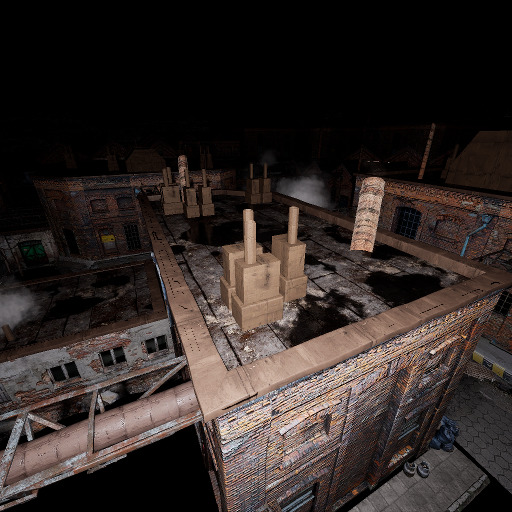}\\
        \teaserimg{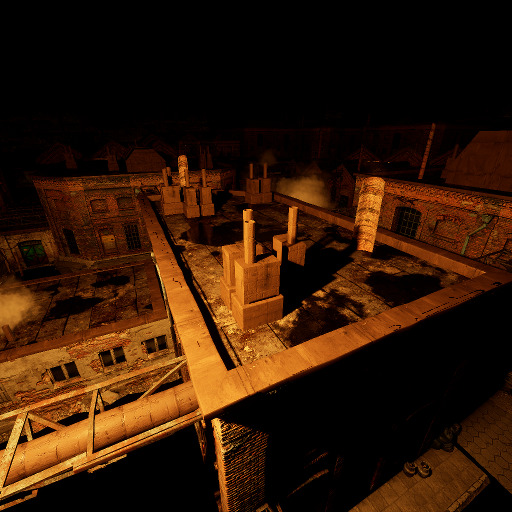}&
        \teaserimg{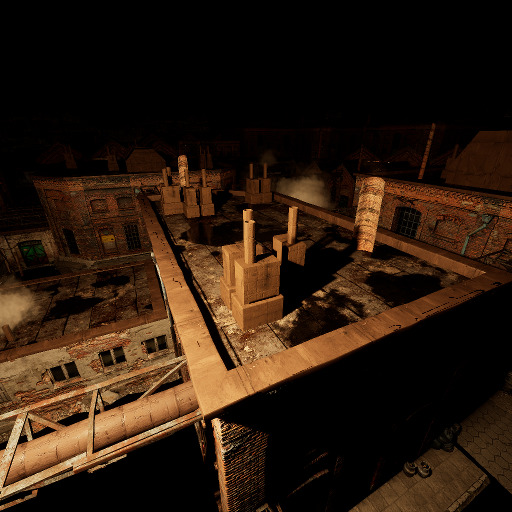}&
        \teaserimg{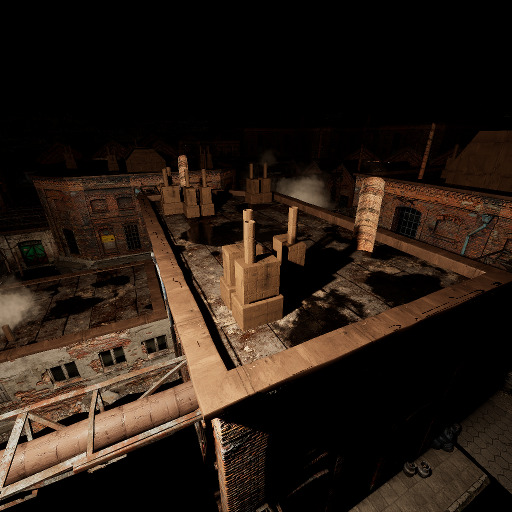}&
        \teaserimg{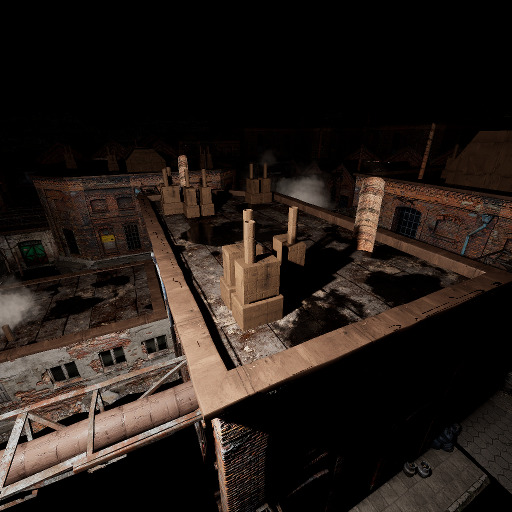}&
        \teaserimg{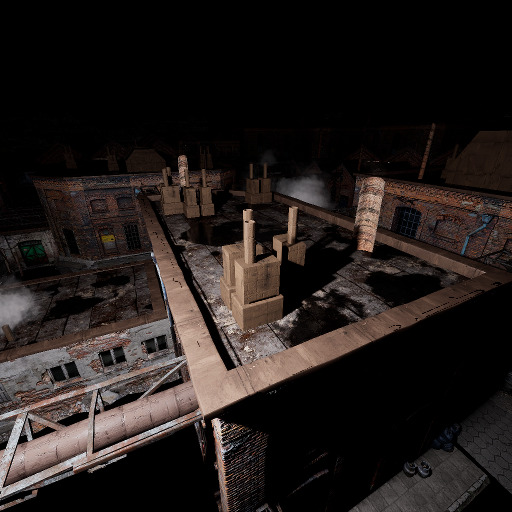}\\
        \teaserimg{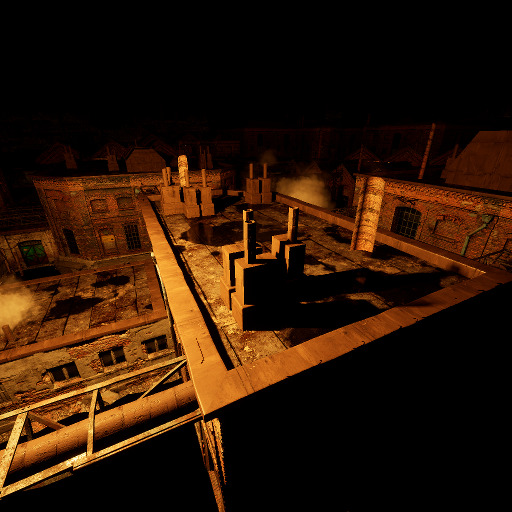}&
        \teaserimg{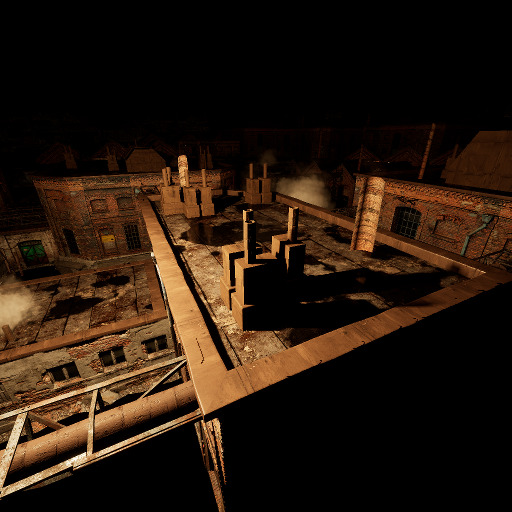}&
        \teaserimg{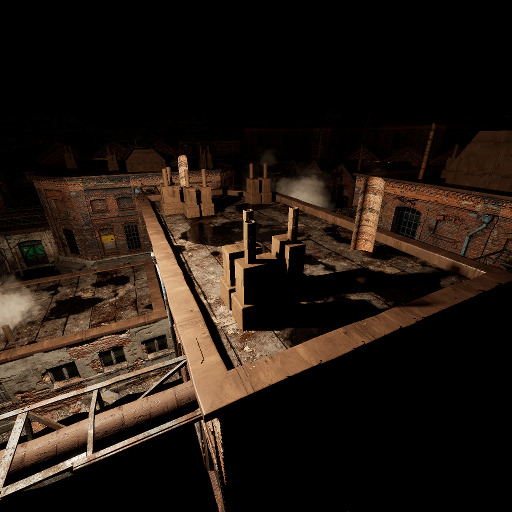}&
        \teaserimg{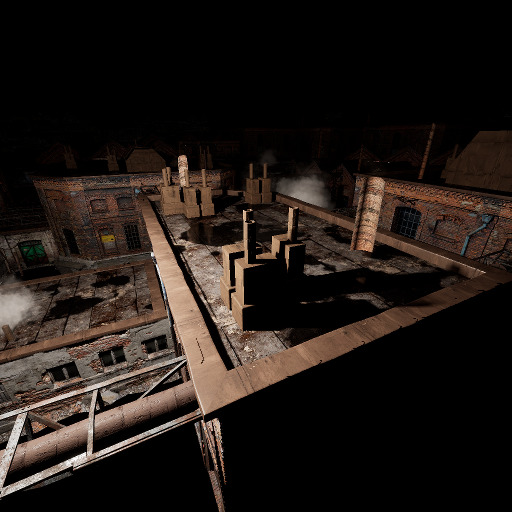}&
        \teaserimg{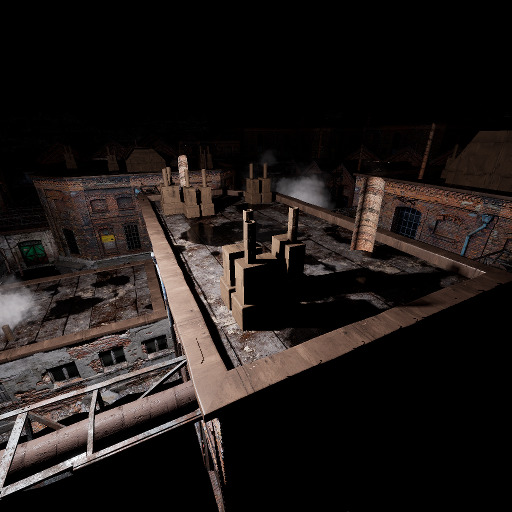}\\
        \teaserimg{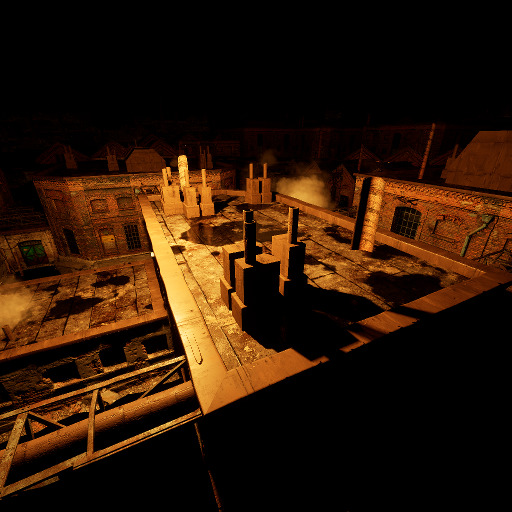}&
        \teaserimg{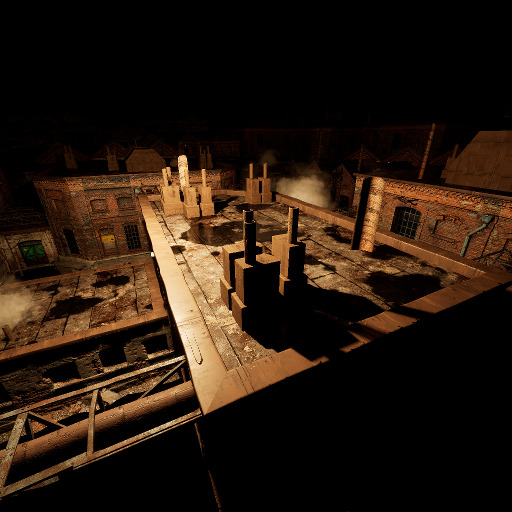}&
        \teaserimg{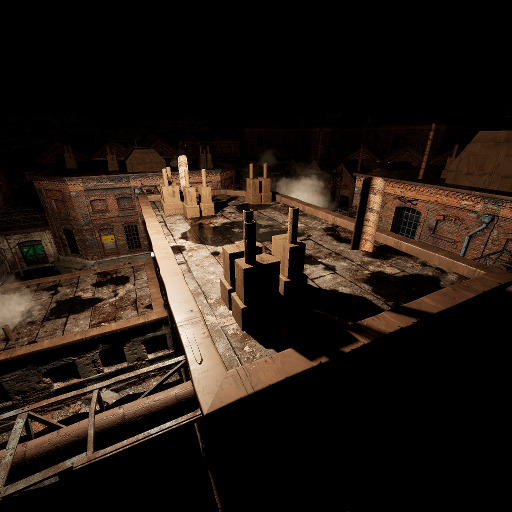}&
        \teaserimg{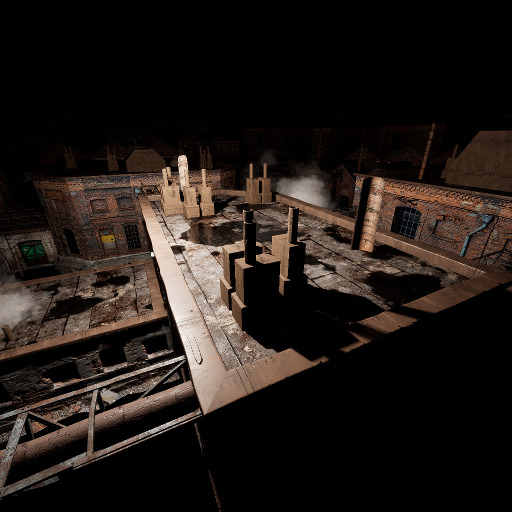}&
        \teaserimg{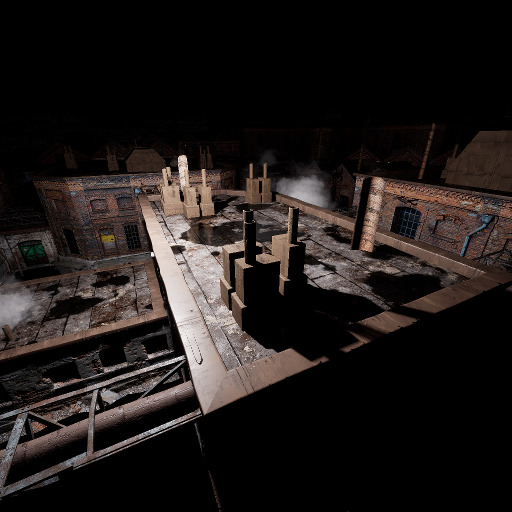}\\
        
    \end{tabu}
    \caption{Example scene captured with 40 different illumination settings. From left to right the illuminant color temperature goes from 2500K to 6500K, and from top to bottom the light source position goes from N to NW.}
    \label{fig:teaser_full}
\end{figure}

\bibliographystyle{splncs04}
\bibliography{egbib}
\end{document}